\documentclass[10pt,twocolumn,letterpaper]{article}

\usepackage{cvpr}
\usepackage{times}
\usepackage{epsfig}
\usepackage{graphicx}
\usepackage{amsmath}
\usepackage{amssymb}
\usepackage{multirow}
\usepackage{array}
\usepackage{booktabs}       % professional-quality tables
\usepackage{latexsym}
\usepackage{arydshln}
\usepackage{bm}
\usepackage{subfigure}
\usepackage{xcolor}
\usepackage{pmat}
\usepackage[misc]{ifsym}
\newcolumntype{M}[1]{>{\centering\arraybackslash}m{#1}}
\usepackage[breaklinks=true,bookmarks=false]{hyperref}

\cvprfinalcopy % *** Uncomment this line for the final submission

\def\cvprPaperID{****} % *** Enter the CVPR Paper ID here

\begin{document}

%%%%%%%%% TITLE
\title{Cross-modality Person re-identification with Shared-Specific Feature Transfer}

\author{Yan Lu$^{1,2}$, Yue Wu$^3$, Bin Liu$^{1,2,\text{\Letter}}$, Tianzhu Zhang$^1$, Baopu Li$^4$, Qi Chu$^{1,2}$, Nenghai Yu$^{1,2}$\\
$^1$School of Information Science and Technology, University of Science and Technology of China, Hefei, China \\
$^2$Key Laboratory of Electromagnetic Space Information, Chinese Academy of Science, Hefei, China \\
$^3$Damo Academy, Alibaba Group, Beijing, China\\
$^4$Baidu Research(USA), 1195 Baudeaux Dr, Sunnyvale, CA, USA\\
\tt\small luyan17@mail.ustc.edu.cn, matthew.wy@alibaba-inc.com, baopuli@baidu.com\\
\tt\small \{flowice, tzzhang, qchu, ynh\}@ustc.edu.cn
% For a paper whose authors are all at the same institution,
% omit the following lines up until the closing ``}''.
% Additional authors and addresses can be added with ``\and'',
% just like the second author.
% To save space, use either the email address or home page, not both
\and
%Yue Wu\\
%Institution2\\
%First line of institution2 address\\
%{\tt\small secondauthor@i2.org}
%\and
}

\maketitle
%\thispagestyle{empty}

%%%%%%%%% ABSTRACT
\begin{abstract}
    Cross-modality person re-identification (cm-ReID) is a challenging but key technology for intelligent video analysis. 
    %Existing works mainly focus on modality-shared representation learning by embedding features of different modalities into a same space. 
    %These methods effectively mine the modality shared features but often ignore the potential of modal-specific cues.
    Existing works mainly focus on learning modality-shared representation by embedding different modalities into a same feature space, lowering the upper bound of feature distinctiveness. 
    %However, only learning the common characteristics means great information loss
    %The proposed algorithm has two parts. The first one is a two-stream feature extractor who can provides modality-shared features and complementary modality-specific features simultaneously. In the second stage, a novel transform network propagates specific and shared features between each inter-modal or intra-modal sample, which can generate features that include more useful information not limited in the modality-shared features. 
    In this paper, we tackle the above limitation by proposing a novel cross-modality shared-specific feature transfer algorithm (termed cm-SSFT) to explore the potential of both the modality-shared information and the modality-specific characteristics.% to boost the re-identification performance. 
    We model the affinities of different modality samples according to the shared features and then transfer both shared and specific features among and across modalities. We also propose a complementary feature learning strategy including modality adaption, project adversarial learning and reconstruction enhancement to learn discriminative and complementary shared and specific features of each modality, respectively.
    The entire cm-SSFT algorithm can be trained in an end-to-end manner. We conducted comprehensive experiments to validate the superiority of the overall algorithm and the effectiveness of each component. The proposed algorithm significantly outperforms state-of-the-arts by 22.5\% and 19.3\% mAP on the two mainstream benchmark datasets SYSU-MM01 and RegDB, respectively.  
    %We model the correspondence between different modalities
    %according to the shared features and then transfer both modality shared and specific
    %features from one modality to another. A channel-wise attention mechanism is
    %proposed to emphasize informative features to further increase the representation ability.
    %Besides, we propose a dual relation learning strategy to jointly learn two modalities in an end-to-end manner. 
\end{abstract}
\begin{figure}[t!]
    \centering
    \includegraphics[width=0.8\linewidth]{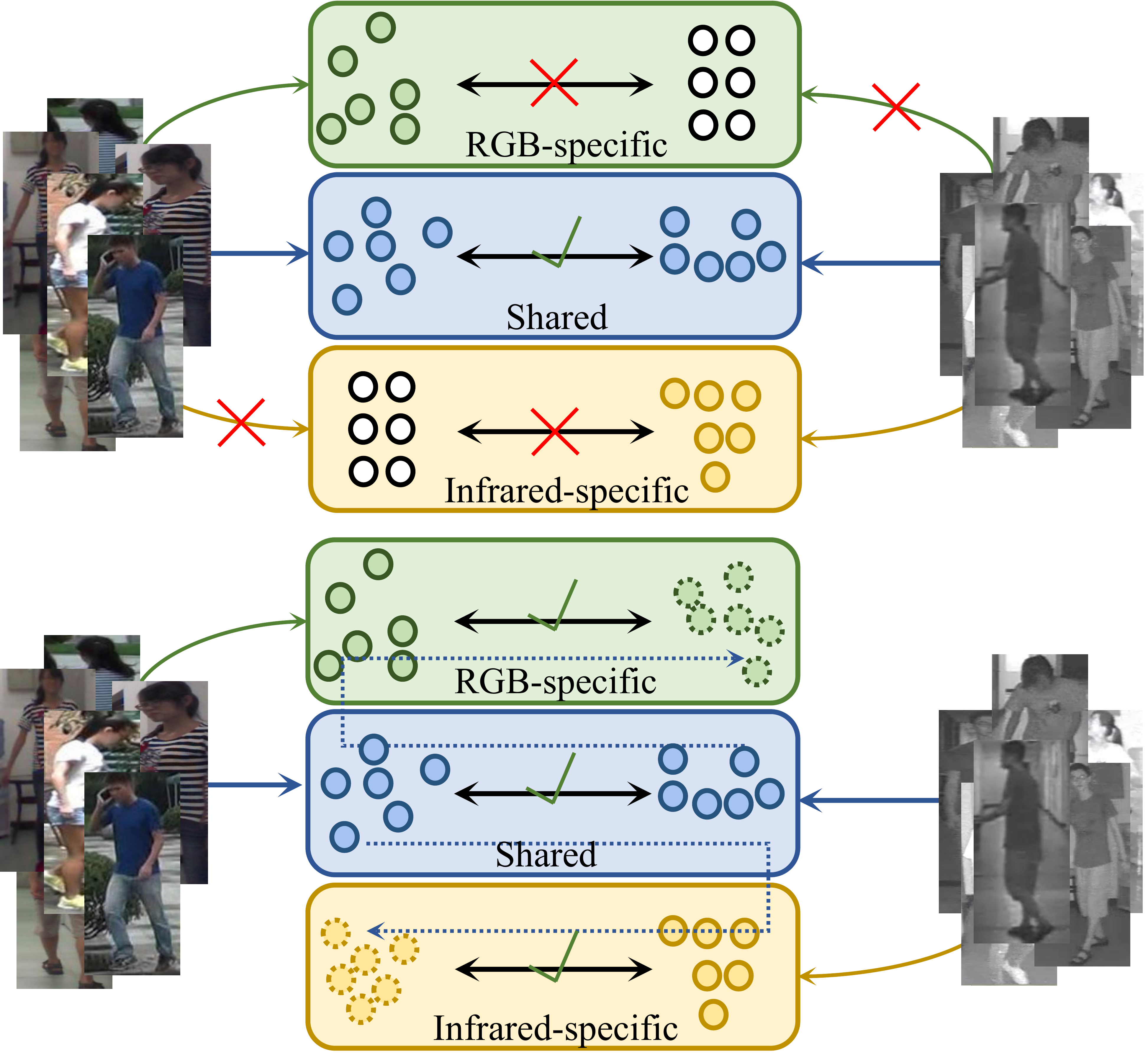}
    \caption{Illustration of the difference between our algorithm and modality-shared feature learning methods. The modality-shared feature learning methods abandon lots of useful specific cues because the modality-specific information cannot be extracted from the other modality. Our algorithm tries to introduce modality-specific features based on the cross-modality near neighbor affinity modeling, effectively utilizing both shared and specific information for each sample.} %% label for entire figure
    \label{fig:motivation}
    %\vspace{-0.3cm}
\end{figure}

%%%%%%%%% BODY TEXT
\footnotetext{Bin Liu is the corresponding author.}
\section{Introduction}
Person re-identification (ReID) aims to find out images of the same person to the query image from a large gallery.
%like cross-camera pedestrian tracking. 
Many works focus on feature learning
~\cite{hermans2017defense,suh2018part}
and metric learning~\cite{chen2015mirror,liao2015person}
on the RGB modality. These methods have achieved great success, especially with the most
recent deep learning technology~\cite{sun2018beyond}.
However, the dependency on bright lighting environments limits their applications
in real complex scenarios. The performance of these methods degrades
dramatically in dark environments where most cameras cannot work well~\cite{wu2017rgb}.
Hence, other kinds of visual sensors like infrared cameras are now widely
used as a complement to RGB cameras to overcome these difficulties, yielding 
popular research interest on RGB-Infrared cross-modality person ReID (cm-ReID).

Compared to conventional ReID task, the major difficulty of cm-ReID is the modality discrepancy resulting from intrinsically distinct imaging processes of different cameras. Some discriminative cues like colors in RGB images are missing in infrared images. Previous methods can be summarized into two major categories to overcome the modality discrepancy: modality-shared feature learning and modality-specific feature compensation. The shared feature learning aims to embed images of whatever modality into a same feature space~\cite{wu2017rgb,ye2018hierarchical,ye2018visible}. The specific information of different modalities such as colors of RGB images and thermal of infrared images are eliminated as redundant information~\cite{dai2018cross}. However, the specific information like colors plays an important role in conventional ReID. With shared cues only, the upper bound of the discrimination ability of the feature representation is limited. As a result, modality-specific feature compensation methods try to make up the missing specific information from one modality to another. Dual-level Discrepancy Reduction Learning (D$^2$RL)~\cite{wang2019learning} is the typical work to generate multi-spectral images to compensate for the lacking specific information by utilizing the generative adversarial network (GAN)~\cite{goodfellow2014generative}. However, a person in the infrared modality can have different colors of clothes in the RGB space. There can be multiple reasonable results for image generation. It's hard to decide which one is the correct target to be generated for re-identification without memorization of the limited gallery set.

In this paper, we tackle the above limitations by proposing a novel cross-modality shared-specific feature transfer algorithm (termed cm-SSFT) to explore the potential of both the modality-shared information and the modality-specific characteristics to boost the re-identification performance. It models the affinities between intra-modality and inter-modality samples and utilizes them to propagate information. Every sample accepts the information from its inter-modality and intra-modality near neighbors and meanwhile shares its own information with them. This scheme can compensate for the lack of specific information and enhance the robustness of the shared feature, thus improving the overall representation ability.
Comparison with the shared feature learning methods are shown in Figure~\ref{fig:motivation}. Our method can exploit the specific information that is unavailable in traditional shared feature learning. Since our method is dependent on the affinity modeling of neighbors, the compensation process can also overcome the choice difficulty of generative methods. Experiments show that the proposed algorithm can significantly outperform state-of-the-arts by 22.5\% and 19.3\% mAP, as well as 19.2\% and 14.4\% Rank-1 accuracy on the two most popular benchmark datasets SYSU-MM01 and RegDB, respectively.

The main contributions of our work are as follows:
\begin{itemize}
    \item We propose an end-to-end cross-modality shared-specific feature transfer (cm-SSFT) algorithm to utilize both the modality shared and specific information, achieving the state-of-the-art cross-modality person ReID performance.
    \item We put forward a feature transfer method by modeling the inter-modality and intra-modality affinity to propagate information among and across modalities according to near neighbors, which can effectively utilize the shared and specific information of each sample.
    \item We provide a novel complementary learning method to extract discriminative and complementary shared and specific features of each modality, respectively, which can further enhance the effectiveness of the cm-SSFT.
\end{itemize}

\if 0
In general, for utilizing specific information, we propose a novel algorithm called Cross-Modality Shared-Specific Feature Transfer for RGB-Infrared person re-identification. We argue that modality-specific information can be further exploited for better performance. The main contributions of our work are as follows:
\begin{itemize}
    \item We propose an end-to-end algorithm to utilize both the modality shared and specific information and achieve the state-of-the-art cross-modality person ReID performance.
    \item For extracting shared features and complementary specific features, we advance a novel disentangle method called Project Adversarial Network (PAN). It can generate shared and complementary specific representations simultaneously.
    \item We put forward a feature transform model called Shared-Specific Transfer Network (SSTN). It can construct the inter- and intra-modality affinity model and propagate information for each sample from its near neighbors, which can compensate for the lacking specific information and enhance the features.  
\end{itemize}
\fi

%The rest of this paper is organized as follows. Section 2 introduces the related
%works. Section 3 illustrates
%the details of the proposed method. Section 4 validates the effectiveness of
%the proposed algorithm with comprehensive experiments.  Section 5 concludes the paper.

%------------------------------------------------------------------------
\section{Related Work}
\textbf{Person ReID. } Person ReID~\cite{zheng2016person} aims to search target person images in a large gallery set with a query image. The recent works are mainly based on
deep learning for more discriminative features~\cite{fang2019bilinear,hou2019interaction,xia2019second,zhou2019discriminative}.
Some of them treat it as a partial feature learning task and pay much attention
to more powerful network structures to better discover, align, and depict the body parts~\cite{guo2019beyond,sun2019perceive,sun2018beyond,liu2015spatio}.
Other methods are based on metric learning, focusing on proper loss functions, like the contrastive loss~\cite{rama16siamese}, triplet loss~\cite{hermans2017defense},
quadruplet loss~\cite{chen2017beyond}, etc. Both kinds of methods try to discard the unrelated cues, such as pose, viewpoint and illumination changing out of the features and the metric space. Recent disentangle based methods extend along this direction further by splitting each sample to identity-related and identitiy-unrelated features, obtaining purer representations without redundant cues~\cite{ham2019learning,zheng2019joint}. 

The aforementioned methods process each sample independently, ignoring the connections between person images. Recent self-attention~\cite{vaswani2017attention,luo2018spectral} and graph-based methods ~\cite{bai2017scalable,shen2018deep,shen2018person,wu2019unsupervised} tried to model the relationship between sample pairs.
Luo~\emph{et al.} proposed the spectral feature transformation method to fuse features between different identities~\cite{luo2018spectral}.
Shen~\emph{et al.} proposed a similarity guided graph neural network~\cite{shen2018person}
and deep group-shuffling random walk~\cite{shen2018deep} to fuse the residual features of different samples to obtain more robust representation. Liu~\emph{et al.} utilized the near neighbors to tackle the unsupervised ReID~\cite{liu2017stepwise}.
%Though explored, cross sample analysis is rarely studied in cross-modality person ReID.

\textbf{Cross-modality matching.}
Cross-modality matching aims to match samples from different modalities, such as cross-modality retrieval~\cite{gu2018look,he2019new,lee2018stacked,liu2019mtfh} and cross-modality tracking~\cite{zhu2018fanet}. 
Cross-modality retrieval has been widely studied for heterogeneous face recognition~\cite{he2018wasserstein} and text-to-image retrieval~\cite{gu2018look,he2019new,lee2018stacked,li2018self,liu2019mtfh}.
~\cite{he2018wasserstein} proposed a two-stream based deep invariant feature representation learning method for heterogeneous face recognition.

\begin{figure*}[!t]
    \centering
    \includegraphics[width=0.95\linewidth]{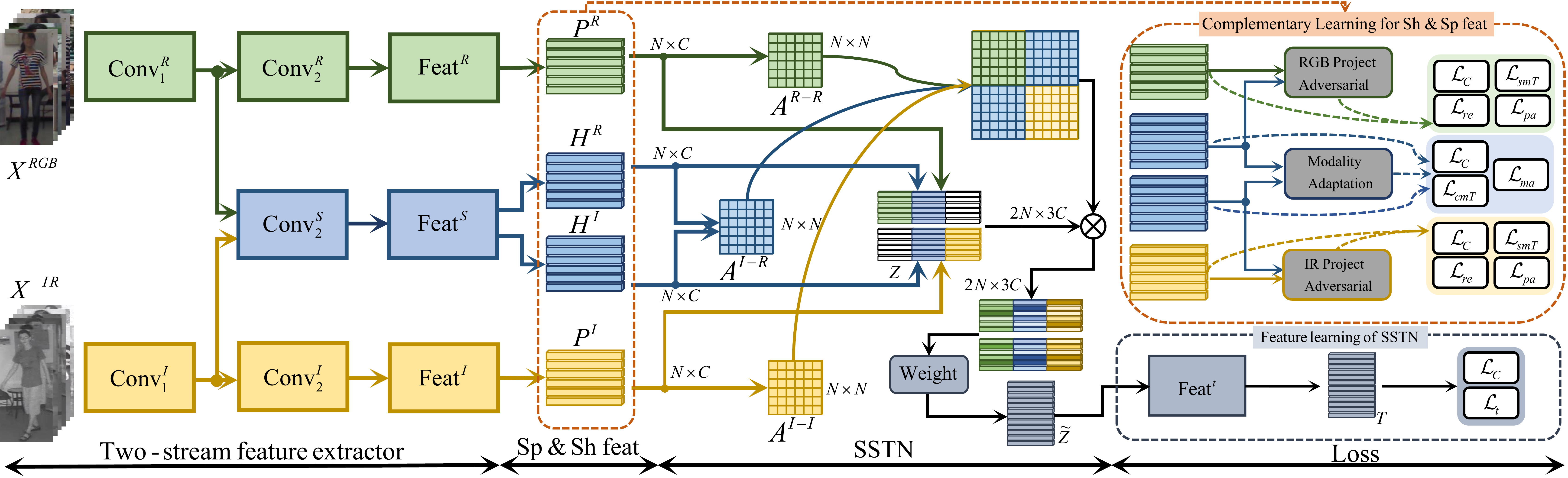}
    \caption{Framework of the cross-modality shared-specific feature transfer algorithm. } %% label for entire figure
    \label{fig:Overview}
    %\vspace{-0.3cm}
\end{figure*}

\textbf{Cross-modality person ReID.}
Cross-modality person ReID aims to match queries of one modality against a gallery set of another modality~\cite{wang2019beyond}, %It has been widely
%studied for heterogeneous face recognition~\cite{he2018wasserstein} and
%text-to-image retrieval~\cite{li2018self}. 
such as text-image ReID~\cite{li2017person,niu2019improving,sarafianos2019adversarial}, RGB-Depth ReID~\cite{hafner2018cross,wu2017robust} and
RGB-Infrared (RGB-IR) ReID~\cite{dai2018cross,feng2019learning,hao2019hsme,kang2019person,kniaz2018thermalgan,lin2019hpiln,wang2019rgb,wang2020cross,wang2019learning,wu2017rgb,ye2018hierarchical,ye2018visible,zhang2019attend}. 
Wu~\emph{et al.} built the largest SYSU-MM01 dataset for RGB-IR person ReID evaluation~\cite{wu2017rgb}.
Ye~\emph{et al.} advanced a two-stream based model and bi-directional top-ranking loss
function for the shared feature embedding~\cite{ye2018hierarchical,ye2018visible}.
To make the shared features purer, Dai~\emph{et al.} suggested a generative adversarial training method for the shared feature learning~\cite{dai2018cross}. 
These methods only concentrate on the shared feature learning and ignore the potential values of specific features. Accordingly, some other works try to utilize modality-specific features and focus on cross-modality GAN.
Kniaz~\emph{et al.} proposed ThermalGAN to transfer RGB images to IR
images and extracted features in IR domain~\cite{kniaz2018thermalgan}. Wang~\emph{et al.} put forward dual-level discrepancy reduction learning based on a bi-directional cycle GAN to reduce the gap between different modalities \cite{wang2019learning}. 
More recently, Wang~\emph{et al.}~\cite{wang2019rgb} constructed a novel GAN model with the joint pixel-level and feature-level constraint, which achieved the state-of-the-art performance. However, it is hard to decide which one is the correct target to be generated from the multiple reasonable choices for ReID.

%Existing feature learning based cm-ReID methods mainly concentrate on shared feature learning or only enhance the discrimination by joint shared and specific feature learning. They all ignored the potential values of the specific features. However, we argue that the modality specific information are always helpful to the cm-ReID and may produce better performance. 

%-------------------------------------------------------------------------

\section{Cross-Modality Shared-Specific Feature Transfer}
%In this section, we illustrate the details of our proposed cross-modality
%shared-specific feature transfer algorithm. 
The framework of the proposed cross-modality shared-specific feature transfer algorithm (cm-SSFT) is shown in Figure~\ref{fig:Overview}. Input images are first fed into the two-stream feature extractor to obtain the shared and specific features. Then the shared-specific transfer network (SSTN) models the intra-modality and inter-modality affinities. It then propagates the shared and specific features across modalities to compensate for the lacked specific information and enhance the shared features. To obtain discriminative and complementary shared and specific features, two project adversarial and reconstruction blocks and one modality-adaptation module are added on the feature extractor. The overall algorithm is trained in an end-to-end manner. 
%The joint feature learning task with the adversarial min-max game may guide more discriminative feature representations.

To better illustrate how the proposed algorithm works, we distinguish the RGB modality, infrared modality and shared space with $R$, $I$ and $S$ in superscript. We use $H$ and $P$ to denote sHared and sPecific features, respectively.

\subsection{Two-stream feature extractor}
%To split each sample to a shared and a complementary specific feature, we propose a disentangle model PAN. This model has two streams, one for RGB and the other one for infrared. The overall pipeline of PAN is shown in Figure~\ref{fig:PAN}. Each sample is fed into the corresponding stream and get a shared and a specific feature map. For the feature learning task, these two kinds of feature maps are pooled and embedded by the corresponding feature block, and then shared and specific feature learning loss functions will add on these features. Except that, to make the shared features purer and specific features more complementary, a modality adaptation and two project adversarial modules are added on shared and specific features of two modalities. 

As shown in Figure~\ref{fig:Overview}, our two-stream feature extractor includes the modality-shared stream (in blue blocks) and the modality-specific stream (green blocks for RGB and yellow blocks for IR). Each input image $X^m$ ($m\in \{R,I\}$) will pass the convolutional layers and the feature blocks to generate the shared feature and specific feature. For better performances, we separate the shared and specific stream at the shallow convolutional layers instead of the deeper fully-connected layers~\cite{ye2018hierarchical}:
\begin{small}\begin{equation}
\begin{aligned}
    H^m = \text{Feat}^S(\text{Conv}_2^S(\text{Conv}_1^m(X^m))),\\
    P^m = \text{Feat}^m(\text{Conv}_2^m(\text{Conv}_1^m(X^m))).
\end{aligned}
\end{equation}\end{small}
To make sure that the two kinds of features are both discriminative, we add the classification loss $\mathcal{L}_c$ on each kind of features respectively:
\begin{small}\begin{equation}
\begin{aligned}
    \mathcal{L}_{c}(H^m) = \mathbb{E}_{i,m}[-\log(p(y_i^m|H_i^{m}))],\\
    \mathcal{L}_{c}(P^m) = \mathbb{E}_{i,m}[-\log(p(y_i^m|P_i^{m}))],
\end{aligned}
\end{equation}\end{small}
where $p(y_i^m|*)$ is the predicted probability of belonging to the ground-truth class $y_i^m$ for the input image $X^m$. The classification loss ensures that features can distinguish the identities of the inputs. Besides, we add a single modality triplet loss ($\mathcal{L}_{smT}$)~\cite{hermans2017defense} on specific  features and a cross-modality triplet loss ($\mathcal{L}_{cmT}$)~\cite{dai2018cross,ye2018visible} on shared features for better discriminability:
\begin{small}
\begin{small}\begin{equation}
    \begin{aligned}
        \mathcal{L}_{smT}(P) &= \sum_{i,j,k}\max[\rho_2+||P_i^{R}-P_j^{R}||-||P_i^{R}-P_k^{R}||,0]\\
        &+ \sum_{i,j,k}\max[\rho_2+||P_i^{I}-P_j^{I}||-||P_i^{I}-P_k^{I}||,0],
    \end{aligned}
\end{equation}\end{small}
\end{small}
\begin{small}
\begin{small}\begin{equation}
    \begin{aligned}
        \mathcal{L}_{cmT}(H) &= \sum_{i,j,k}\max[\rho_1+||H_i^{R}-H_j^{I}||-||H_i^{R}-H_k^{I}||,0]\\
        &+\sum_{i,j,k}\max[\rho_1+||H_i^{I}-H_j^{R}||-||H_i^{I}-H_k^{R}||,0],
    \end{aligned}
\end{equation}\end{small}
\end{small}
where $\rho_1$ and $\rho_2$ are the margins of $\mathcal{L}_{cmT}$ and  $\mathcal{L}_{smT}$, respectively. $i$, $j$, $k$ represent indices of the anchor, positive of the anchor and negative of the anchor of triplet loss $(y_i=y_j,y_i \not= y_k)$.

%\textbf{Optimization of PAN.} 
%The loss functions of PAN can be rewritten as follow:
%  \begin{small}\begin{equation}
%  \begin{aligned}
%      &l_{PAN_G} = l^{sh}_{id}+\lambda _1 l^{sp}_{id}+\lambda _2 l_{cmT}+\lambda _3 l_{smT}+\lambda _4 l_{re}\\
%      &l_{PAN_D} = \lambda _5 l_{ma}+\lambda _6 l_{pa}
%  \end{aligned}
%  \end{equation}\end{small}  
%where $l_{PAN_G}$ means the loss of the generation steps. Minimizing $l_{PAN_G}$ means do the feature learning and reconstruction task together. The $l_{PAN_D}$ is the loss of discrimination steps. Maximize this loss can %guide the model to extract pure shared and complementary specific features. The two steps are as follow:
%  \begin{small}\begin{equation}
%  \begin{aligned}
%      \mathop{\arg\min}_{\theta_b} (l_{PAN_G}-l_{PAN_D}).
%  \end{aligned}
%  \end{equation}\end{small}  

\subsection{Shared-Specific Transfer Network}
The two-stream network extracts the shared and specific features for each modality. 
For unified feature representation, we pad and denote the features of each modality with a three-segment format: [RGB-specific; shared; Infrared-specific] as follows:
\begin{small}\begin{equation}
\begin{aligned}
    Z^R_i=[P_i^{R};H_i^{R};\mathbf{0}],  \quad
    Z^{I}_i=[\mathbf{0};H_j^{I};P_j^{I}].
\end{aligned}
\end{equation}\end{small}
Here, $\mathbf{0}$ denotes the padding zero vector, which means that samples of the RGB modality have no specific features of infrared modality, and vice versa. $[\bullet;\bullet]$ means concatenation in the columan dimension. For cross-modality retrieval, we need to transfer the specific features from one modality to another to compensate for these zero-padding vectors.
Motivated by graph convolutional network (GCN), we utilize the near neighbors to propagate information and meanwhile maintain the context structure of the overall sample space. The proposed shared-specific transfer network can make up the lacking specific features and enhance the robustness of the overall representation jointly. As shown in Figure~\ref{fig:Overview}, SSTN first models the affinity of samples according to the two kinds of features. Then it propagates both intra-modality and inter-modality information with the affinity model. Finally, the feature learning stage guides the optimization of the whole process with classification and triplet losses. 

%After extracting shared and specific features, it is still impossible to use the specific feature directly because we cannot extract $m$ modality-specific features from the $m'$ modality samples ($m\not= m'$. When $m$=rgb, $m'$=infrared. Similarly, when $m$=infrared, $m'$=RGB.). An intuitive solution is to utilize information of the cross-modality near neighbors. The graph convolutional network (GCN) is suitable for this motivation because the GCN model may not only propagate information between each sample pairs but also can maintain the context structure of the overall sample space, which may provide better performance. Motivated by this, we propose a novel GCN based Shared-Specific Transfer Network (SSTN), as shown in Figure~\ref{fig:Overview}, to construct the lacking specific features and enhance the robustness of the overall representation jointly. The SSTN first constructs an affinity model based on the two kinds of features and then propagated information based on the affinity model. Finally, a feature block embed the output of SSTN to the final feature space. The pipeline of the SSTN can be described with the following equation:  

\textbf{Affinity modeling.} 
We use the shared and specific features to model the pair-wise affinity. We take the specific features to compute the intra-modality affinity and the shared features for inter-modality as follows:
\begin{small}\begin{equation}
\label{equ:sim}
%\begin{aligned}
    A^{m,m}_{ij}=d(P^{m}_i,P^{m}_j), \quad  A^{m,m'}_{ij}=d(H^{m}_i,H^{m'}_j), \\
%\end{aligned}\label{equ:intra-sim}
\end{equation}\end{small}
where $A^{m,m}_{ij}$ is the intra-modality affinity between the $i$-th sample and the $j$-th sample, both of which belong to the $m$ modality. $A^{m,m'}_{ij}$ is the inter-affinity.
%The inter-modality similarities are calculated as:
%\begin{small}\begin{equation}
%\begin{aligned}
%   .\\
%\end{aligned}\label{equ:inter-sim}
%\end{equation}\end{small}
$d(a,b)$ is the normalized euclidean distance metric function:
\begin{small}\begin{equation}
d(a,b)=1-0.5\cdot \left \| \frac{a}{\left \| a \right \|}-\frac{b}{\left \| b\right \|} \right \|.
\end{equation}\end{small}
The intra-similarity and inter-similarity represent the relation between each sample with others of both the same and different modalities. We define the final affinity matrix as: 
\begin{small}\begin{equation}
    A=\begin{pmat}[{|}]
            \mathcal{T}(A^{R,R},k) & \mathcal{T}(A^{R,I},k) \cr\-
            \mathcal{T}(A^{I,R},k) & \mathcal{T}(A^{I,I},k) \cr
    \end{pmat},
    \label{eq:affinity}
\end{equation}\end{small}
where $\mathcal{T}(\bullet,k)$ is the near neighbor chosen function. It keeps the top-$k$ values for each row of a matrix and sets the others to zero.

\textbf{Shared and specific information propagation.} 
The affinity matrix represents the similarities across samples. SSTN utilizes this matrix to propagate features. Before this, features of the RGB and infrared modalities are concatenated in the row dimension, each row of which stores a feature of a sample:

\begin{small}\begin{equation}
\label{normalize}
\begin{aligned}
  Z=\begin{pmat}[{}]
            Z^R \cr\-
            Z^I \cr
    \end{pmat}.
\end{aligned}
\end{equation}\end{small}

Following the GCN approach, we obtain the diagonal matrix  $D$ of the affinity matrix $A$ with $d_{ii} = \sum_j{A_{ij}}$.
%To normalize the affinity matrix, we calculate the diagonal matrix $D$ as the following: $d_i = \sum_j{S_{ij}}$. 
The padded features are first propagated with the near neighbor structure ($D^{-\frac{1}{2}}AD^{-\frac{1}{2}}Z$) and then fused by a learnable non-linear transformation. After feature fusion, the propagated features will include shared features and specific features of both the two modalities. The propagated features $\widetilde{Z}$ are calculated as:
\begin{small}\begin{equation}
\label{pipeline}
\begin{aligned}
    \widetilde{Z} = \begin{pmat}[{}]
            \widetilde{Z}^R \cr\-
            \widetilde{Z}^I \cr
    \end{pmat} 
    =\sigma(D^{-\frac{1}{2}}AD^{-\frac{1}{2}}ZW),\\
\end{aligned}
\end{equation}\end{small}
where $\sigma$ is the activation function which is ReLU in our implementation. $W$ is the learnable parameters of SSTN. These propagated features are finally fed into a feature learning stage to optimize the whole learning process. The transferred features $T$ are denoted as:
\begin{small}\begin{equation}
\label{dgzp_feat}
\begin{aligned}
T=\begin{pmat}[{}]
            T^R \cr\-
            T^I \cr
    \end{pmat} 
    =\text{Feat}^{t}(\widetilde{Z}).
\end{aligned}
\end{equation}\end{small}

Following the common feature learning principle, we use the classification loss for feature learning:
\begin{small}\begin{equation}
\begin{aligned}
    \mathcal{L}_c(T^m) = \mathbb{E}_{i,m}[-\log(p(y_i^m|T_i^m))].
\end{aligned}
\end{equation}\end{small}
In additoin, we use the triplet loss on the transferred feature to increase the discrimination ability. Since the transferred features include both shared features and specific features of two modalities. We add both the cm-triplet loss $\mathcal{L}_{cmT}(T)$ and sm-triplet loss $\mathcal{L}_{smT}(T)$ on it for better discrimination:
\begin{small}\begin{equation}
    \begin{aligned}
        \mathcal{L}_t(T) &= \mathcal{L}_{cmT}(T)  + \mathcal{L}_{smT}(T) \\
        &= \sum_{i,j,k}\max[\rho_1+||T_i^{R},T_j^{I}||-||T_i^{R},T_k^{I}||,0]\\
        &+\sum_{i,j,k}\max[\rho_1+||T_i^{I},T_j^{R}||-||T_i^{I},T_k^{R}||,0]\\
        &+ \sum_{i,j,k}\max[\rho_2+||T_i^{R},T_j^{R}||-||T_i^{R},T_k^{R}||,0]\\
        &+\sum_{i,j,k}\max[\rho_2+||T_i^{I},T_j^{I}||-||T_i^{I},T_k^{I}||,0].
    \end{aligned}
\end{equation}\end{small}

\subsection{Shared and specific complementary learning}

SSTN explores a new way to utilize both shared the specific features to generate more discriminative representation. However, the overall performance may still suffer from the information overlap between shared and specific features. Firstly, if shared features contain much modality-specific information, the reliability of the inter-similarity matrix in equation~\eqref{equ:sim} will be affected, leading to inaccurate feature transfer. Secondly, if the specific features are highly related to the shared features, the specific features can only provide little complement to the shared features. The redundant information in the specific features will also affect the sensitivity of the intra-modality similarity matrix in equation~\eqref{equ:sim} due to the shared information. 
To alleviate these two problems, we utilize the modality adaptation~\cite{dai2018cross} to filter out modality-specific information from the shared features. We also propose a project adversarial strategy and reconstruction enhancement for complementary modality-specific feature learning. 

%The SSTN combines the specific features to generate more discriminative features. But there are some bad cases which maybe reduce effectiveness. (1). Shared features include many modality information. This will decrease the discriminative ability of the shared feature, which will drop the reliable of the inter-similarity parts in the affinity matrix. (2). The specific features are highly related to the shared features. There may be a large amount of redundant information between shared and specific features. If in this situation, the specific features will only play a limited role. For example, if the specific features contain all repeated information of the shared features, it may be easy for them to fool the modality discriminator (only need to carry some tags of modality) and converge in the feature learning task (because shared features always include identity information). These kinds of specific features are useless, providing little complementary information for the shared features. To alleviate these two problems, we first utilize modality adaptation~\cite{dai2018cross} to filter out the modality information the shared feature. Then we propose the Project Adversarial (PA) to make the specific features more complementary. 

\textbf{Modality adaptation for shared features.} 
To purify the shared features to be unrelated to modalities, we utilize the modality discriminator~\cite{dai2018cross} with three fully-connected layers to classify the modality of each shared feature:
%To make the shared features purer, an intuitive idea is to push the modality-related out of the shared features. So we utilize the modality discriminator~\cite{dai2018cross} that includes three fully-connected layers to classify the modality of each shared feature:
\begin{small}\begin{equation}
\begin{aligned}
    \mathcal{L}_{ma} &= \mathbb{E}_{i,m}[-\log(p(m|H_i^{m},\Theta_D))],
\end{aligned}
\end{equation}\end{small} 
where $\Theta_D$ represents parameters of the modality discriminator. $p(m|H_i^{m})$ is the predicted probability of feature $H_i^{m}$ belonging to modality $m$. In the discrimination stage, the modality discriminator will try to classify the modality of each shared feature. In the generation stage, the backbone network will generate features to fool the discriminator. This min-max game will make the shared features not contain any modality-related information.

\textbf{Project adversarial learning for specific features.}
To make the specific features uncorrelated with the shared features, we propose the project adversarial strategy. In the training stage, we project the specific features to the shared features of the same sample. The projection error is used as the loss function
  \begin{small}\begin{equation}
  \begin{aligned}
      \mathcal{L}_{pa} &= \mathbb{E}_{i,m}\left [\left \| \Theta^{m}\cdot P_i^m-H_i^m\right \| \right],
  \end{aligned}
  \end{equation}\end{small}  
where $\Theta _p^{m}$ represents the projection matrix for modality $m$. In this equation, "$\cdot$" means matrix multiply. 
Similarly, in the discrimination stage, optimization of $\Theta _p^{m}$ will try to project the specific features to the corresponding shared features. While in the generation stage, the backbone network will generate specific features uncorrelated with shared features to fool the projection. This adversarial training can make the feature spaces of the two kinds of features linearly independent. Alternatively minimizing and maximizing the projection loss will lead the backbone network to learn specific patterns different from shared features.

%When giving a shared feature and a specific feature of one sample, a linear projection projects the specific feature to the shared feature. The projection error is as the loss function:
%  \begin{small}\begin{equation}
%  \begin{aligned}
%      l_{pa} &= \mathbb{E}_{i,m}\left [\left \| \Theta _p^{m}\cdot p_i^{m}-h_i^{m}\right \| \right].
%  \end{aligned}
%  \end{equation}\end{small}  
 %where $\Theta _p^{m}$ represents the project matrix for the modality $m$. In the training stage, the linear projection will project the specific features to the corresponding shared features. The backbone network will escape the linear projection constraint as far as possible. This adversarial training can make the feature spaces of the two kinds of features be linearly independent, reducing the redundancy  between two kinds of features. In this case, the specific feature learning can guide specific features to learn patterns that are different from shared features, which is non-trivial.
 
\textbf{Reconstruction enhancement.} Modality adaption and project adversarial learning make sure that the shared and specific features do not contain correlated information between each other. To enhance both features to be complementary, we use a decoder network after features of each modality to reconstruct the inputs. We concatenate the shared and specific features and feed them to the decoder $\mathcal{D}{e}$:
\begin{small}\begin{equation}
\label{eq:decoder}
\begin{aligned}
    \hat{X}^m&=\mathcal{D}{e}^m([{P}^{m};{H}^{m}]),\\
\end{aligned}
\end{equation}\end{small} 
where $[\bullet;\bullet]$ means feature concatenation. The $L_{2} $ loss is used to evaluate the quality of the reconstructed images:
\begin{small}\begin{equation}
\begin{aligned}
    \mathcal{L}_{re} &= \mathbb{E}_{i,m}[L_{2}(X_i^m,\hat{X}_i^m)].
\end{aligned}
\end{equation}\end{small} 
The reconstruction task makes a constraint on the overall information loss. Combined with project modality adaption and adversarial learning, shared and specific features are guided to be self-discriminate and mutual-complementary.

\subsection{Optimization}
Our proposed algorithm is trained in an end-to-end manner with the adversarial min-max games. We mix the loss function based on the principle that the classification and the triplet share the same importance. So the feature learning losses of each part are as follows:
\begin{small}\begin{equation}
    \begin{aligned}
        &\mathcal{L}(H)=\mathcal{L}_c(H^m)+0.5\cdot \mathcal{L}_{cmT},\\
        &\mathcal{L}(P)=0.5\cdot (\mathcal{L}_c(P^R)+\mathcal{L}_c(P^I))+0.5\cdot \mathcal{L}_{smT},\\
        &\mathcal{L}(T)=\mathcal{L}_c(T)+0.25\cdot \mathcal{L}_t(T).
    \end{aligned}
\end{equation}\end{small}
Furthermore, we think that the backbone feature extractor and SSTN share the same importance. Hence, the overall feature learning loss is as follows:
\begin{small}\begin{equation}
    \begin{aligned}
        \mathcal{L}_{feat}=\mathcal{L}(H)+\mathcal{L}(P)+\mathcal{L}(T).
    \end{aligned}
\end{equation}\end{small}
Therefore, the overall loss functions of the min and the max steps of each part are as follows:
\begin{small}\begin{equation}
    \begin{aligned}
        \mathcal{L}_{min}=&\mathcal{L}_{feat}+\lambda_1\mathcal{L}_{re}-\lambda_2\mathcal{L}_{ma}-\lambda_3\mathcal{L}_{pa},\\
        \mathcal{L}_{max}=&-\lambda_2\mathcal{L}_{ma}-\lambda_3\mathcal{L}_{pa}.
    \end{aligned}
\end{equation}\end{small}
The optimization process includes two sub-processes: (1) fix each discriminator and minimize $\mathcal{L}_{min}$. (2) fix all modules excluding the three discriminators and  maximize the $\mathcal{L}_{max}$. Support $\Theta_N$ denotes the parameters of the overall networks except all the other discriminators. The alternative learning process is:
\begin{small}\begin{equation}
    \begin{aligned}
        \hat{\Theta}_N=\mathop{\arg\min}_{\Theta_N}\mathcal{L}_{min}(\Theta_N,\hat{\Theta}_D,\hat{\Theta}^{m}),\\
        \hat{\Theta}_D,\hat{\Theta}^{m}=\mathop{\arg\max}_{\Theta_D, \Theta^m}\mathcal{L}_{max}(\hat{\Theta_N},\Theta_D,\Theta^{m}).
    \end{aligned}
\end{equation}\end{small}

\begin{table*}[!t]
    \centering
    \fontsize{8}{8}\selectfont
    \caption{Comparison on SYSU-MM01. r1, r10, r20 denote Rank-1, 10, 20 accuracies (\%)
    .}
    \label{tab:sysu_comp}
    \begin{tabular}{M{3cm} M{0.3cm} M{0.3cm} M{0.3cm} M{0.3cm} M{0.3cm} M{0.3cm} M{0.3cm} M{0.3cm} M{0.3cm} M{0.3cm} M{0.3cm} M{0.3cm} M{0.3cm} M{0.3cm} M{0.3cm} M{0.3cm} M{0.3cm}}
        \hline
        \multirow{3}{*}{Method}&
        \multicolumn{8}{c}{All-search}&\multicolumn{8}{c}{Indoor-search}\cr\cline{2-17} &
        \multicolumn{4}{c}{Single-shot}&\multicolumn{4}{c}{Multi-shot}&
        \multicolumn{4}{c}{Single-shot}&\multicolumn{4}{c}{Multi-shot}\cr\cline{2-17} &
        r1&r10&r20&mAP&r1&r10&r20&mAP&r1&r10&r20&mAP&r1&r10&r20&mAP\cr\hline
        HOG\cite{dalal2005histograms}&2.76&18.3&31.9&4.24&3.82&22.8&37.6&2.16&3.22&24.7&44.5&7.25&4.75&29.2&49.4&3.51\cr
        LOMO\cite{liao2015person}&3.64&23.2&37.3&4.53&4.70&28.2&43.1&2.28&5.75&34.4&54.9&10.2&7.36&40.4&60.3&5.64\cr
        Zero-Padding\cite{wu2017rgb}&14.8&54.1&71.3&15.9&19.1&61.4&78.4&10.9&20.6&68.4&85.8&26.9&24.4&75.9&91.3&18.6\cr
        TONE+HCML\cite{ye2018hierarchical}&14.3&53.2&69.2&16.2 &-&-&-&-&-&-&-&-&-&-&-&-\cr
        BDTR\cite{ye2018visible}&17.0&55.4&72.0&19.7&-&-&-&-&-&-&-&-&-&-&-&-\cr
        D-HSME\cite{hao2019hsme}&20.7&62.8&78.0&23.2&-&-&-&-&-&-&-&-&-&-&-&-\cr
        IPVT+MSR\cite{kang2019person}&23.2&51.2&61.7&22.5&-&-&-&-&-&-&-&-&-&-&-&-\cr
        cmGAN\cite{dai2018cross}&27.0&67.5&80.6&27.8&31.5&72.7&85.0&22.3&31.6&77.2&89.2&42.2&37.0&80.9&92.1&32.8\cr
        D$^2$RL\cite{wang2019learning}&28.9&70.6&82.4&29.2&-&-&-&-&-&-&-&-&-&-&-&-\cr
        DGD+MSR\cite{feng2019learning}&37.4&83.4&93.3&38.1&43.9&86.9&95.7&30.5&39.6&89.3&97.7&50.9&46.6&93.6&98.8&40.1\cr
        JSIA-ReID\cite{wang2020cross}&38.1&80.7&89.9&36.9&45.1&85.7&93.8&29.5&43.8&86.2&94.2&52.9 &52.7&91.1&96.4&42.7\cr
        AlignGAN\cite{wang2019rgb}&42.4&85.0&93.7&40.7&51.5&89.4&95.7&33.9&45.9&87.6&94.4&54.3&57.1&92.7&97.4&45.3\cr\hline
%ours&{\bf 35.77}&{\bf 77.55}&{\bf 89.68}&{\bf 42.75}&{\bf39.76}&{\bf 79.09}&{\bf 90.98}&{\bf 41.30}
        cm-SSFT (Ours)&{\bf 61.6}&{\bf 89.2}&{\bf 93.9}&{\bf 63.2}&{\bf63.4}&{\bf 91.2}&{\bf 95.7}&{\bf 62.0}&{\bf 70.5}&{\bf 94.9}&{\bf 97.7}&{\bf 72.6}&{\bf  73.0}&{\bf 96.3}&{\bf 99.1}&{\bf 72.4}\cr\hline
    \end{tabular}
    %\vspace{-0.3cm}
\end{table*}

In order to ensure the training effectiveness, every batch contains the equal number of RGB and infrared samples. The details of the sampling strategy are introduced in the implementation details. In the test stage, we utilize the two-stream network to extract disentangled features from the RGB set and the infrared set. We use SSTN to transfer modality-shared and modality-specific features. All features are $L_2$-normalized and we use the Euclidean distance to compute the final ReID performance.

\section{Experiments}
In this section, we conduct comprehensive experiments to validate the
effectiveness of the proposed cross-modality shared-specific feature transfer algorithm as well as each of its components.

\subsection{Experimental settings}
\textbf{Datasets.}
SYSU-MM01 is a large-scale and frequently used RGB-IR cross-modality ReID
dataset~\cite{wu2017rgb}. Images are collected from four RGB cameras and two IR
cameras, in both indoor and outdoor environments.
The training set contains 395 persons, with 22,258 RGB images and 11,909 IR
images. The test set contains 96 persons, with 3,803 IR images for query and 301/3010 (one-shot/multi-shot) randomly selected
RGB images as the gallery. There are two accordingly evaluation
modes for RGB-IR ReID: \emph{indoor-search} and \emph{all-search}~\cite{wu2017rgb}. RegDB is collected by dual camera systems~\cite{sensors17}. There are 412 identities and 8,240 images in total, with 206 identities for training and 206 identities for testing. Each identity has 10 different thermal (IR) images and 10 different visible (RGB) images. There are also two evaluation modes. One is Visible to Thermal to search IR images from a Visible image. The other mode is Thermal to Visible to search RGB images from a infrared image. This dataset has 10 trials with different splits of the dataset. We evaluate our model on the 10 trials to achieve statistically stable results.

\textbf{Evaluation protocols.}
%Following the standard evaluation protocols in~\cite{wu2017rgb}, we adopt the
%cumulative matching characteristics (CMC) curve and mean average
%precision (mAP) as evaluation metrics.
All the experiments follow the standard evaluation protocol in existing RGB-IR cross-modality ReID methods. Queries and galleries images are from different modalities. And then, the standard cumulated matching characteristics (CMC) curve and mean average precision (mAP) are adopted. %Following the settings proposed in ~\cite{wu2017rgb}, the gallery set is randomly sampled from the whole testing set. The above protocol is run 10 times to get the average performance. 

\textbf{Implementation details.}
We use Resnet50~\cite{he16resnet} as the backbone network, with the first convolutional layer, the 1st and 2nd bottlenecks as $\text{Conv}_1$. $\text{Conv}_2$ is the 3rd and 4th bottlenecks. $k$ in Eq.~\eqref{eq:affinity} is set to 4. $\lambda_1$, $\lambda_2$ and $\lambda_3$ are set to 1.0, 0.2 and 0.2, respectively. We change the stride of the last convolutional layer in the backbone to 1 to benefit the learning of reconstruction decoders which are composed of 4 sub-pixel convolutional layers with channels all set to 64~\cite{shi2016real}. We adopt the data and network augmentation methods in BoT for ReID ~\cite{luo2019bag} to enhance the performance. For fairness, we also give out results without any augmentation. The augmentations include: (1) the feature blocks are all set to BNNeck~\cite{luo2019bag}; (2) the input images are augmented with random erasing~\cite{zhong2017random}. The whole algorithm is optimized with Adam for 120 epochs with a batch size of 64 and a learning rate of $0.00035$, decaying 10 times at $40$, $70$ epoch. Each mini-batch is comprised of 8 identities with 4 RGB images and 4 infrared images for each identity.

%The backbone network ($\text{Conv}_m^1$+$\text{Conv}_m^2/\text{Conv}_{\text{sh}}^2$) is set to Resnet50 ~\cite{he16resnet} (All $\text{Conv}^1$ are set to the the first convolutional layer, the 1st and 2nd bottleneck of ResNet50. $\text{Conv}^2$ are all set to the 3rd and 4-th bottleneck of ResNet50. ). $k$ in Eq.~\eqref{eq:affinity} is set to 4. $\lambda_1$, $\lambda_2$ and $\lambda_3$are set to 1.0, 0.2 and 0.2,  respectively. We set the stride of the last convolutional layer in the backbone to 1, which can benefit the learning of decoders. The decoders are composed of 4 sub-pixel convolutional layers~\cite{shi2016real} whose channels are all set to 64.  Besides, we use some tricks from BoT for ReID~\cite{luo2019bag} to enhance the performance. For fairness, all the ablation study are run in this setting and we always give the results without any trick. The tricks are as follows: (1). The Feature blocks are all set as BNNeck~\cite{luo2019bag}. (2). The image are augmented with the random erasing~\cite{zhong2017random}. The whole algorithm is optimized with Adam for 120 epochs with a batch size of 64 and a learning rate of $0.00035$, decaying 10 times at $40$, $70$ epoch. Each mini-batch is comprised of 8 identities and 4 RGB images and 4 infrared images for each identity.

\subsection{Comparison with state-of-the-art methods. }
In this subsection, we compare our proposed algorithm with the baselines as
well as the state-of-the-art methods, including Zero-Padding~\cite{wu2017rgb},
TONE~\cite{ye2018hierarchical}, BDTR~\cite{ye2018visible}, cmGAN~\cite{dai2018cross}, D$^2$RL\cite{wang2019learning}, MSR\cite{feng2019learning}, D-HSME\cite{hao2019hsme}, IPVT\cite{kang2019person}, JSIA-ReID\cite{wang2020cross} and AlignGAN\cite{wang2019rgb}.

The results on SYSU-MM01 are shown in Table~\ref{tab:sysu_comp}. 
The proposed algorithm outperforms other methods by a large margin. Specifically,
in all-search mode, our method surpasses AlignGAN by 19.2\% on
Rank-1 accuracy and 22.5\% on mAP in the single-shot setting. The multi-shot setting
exhibits a similar phenomenon. Compared with single-shot evaluation, mAP of most other methods drop significantly by about 5\% or even more. But our method only drops 1.2\%. 
This validates that the features extracted by our algorithm are much more
discriminative, which can provide higher recall than other methods when the gallery size increases. For indoor-search mode, our method also gets
the best performance on all the evaluation metrics, demonstrating the robustness of the proposed algorithm.

The results on RegDB are shown in~\ref{tab:regdb_comp}. Our method always suppresses others by a large margin. For the Visible to Thermal mode, our method surpasses the state-of-the-art method by 14.4\% on Rank-1 and 19.3\% on mAP. For Thermal to Visible, the advantages are 14.7\% on Rank-1 and 18.3\% on mAP.

\begin{table}[!t]
    \centering
    \fontsize{8}{8}\selectfont
    \caption{Comparison on RegDB.}
    \label{tab:regdb_comp}
    \begin{tabular}{M{3cm} M{0.7cm} M{0.7cm} M{0.7cm} M{0.7cm}}
        \hline
        \multirow{2}{*}{Method}&
        \multicolumn{2}{c}{\emph{Visible to Thermal}}&\multicolumn{2}{c}{\emph{Thermal to Visible}}\cr\cline{2-5}&
        r1&mAP&r1&mAP\cr\hline
        HOG\cite{dalal2005histograms}&13.5&10.3&-&-\cr
        LOMO\cite{liao2015person}&0.80&2.28&-&-\cr
        Zero-Padding\cite{wu2017rgb}&17.8&18.9&16.7&17.9\cr
        TONE+HCML\cite{ye2018hierarchical}&24.4&20.8&21.7&22.2\cr
        BDTR\cite{ye2018visible}&33.5&31.8&32.7&31.1\cr
        D$^2$RL\cite{wang2019learning}&43.4&44.1&-&-\cr
        DGD+MSR\cite{feng2019learning}&48.4&48.7&-&-\cr
        JSIA-ReID\cite{wang2020cross}&48.5&49.3&48.1&48.9\cr
        D-HSME\cite{hao2019hsme}&50.9&47.0&50.2&46.2\cr
        IPVT+MSR\cite{kang2019person}&58.8&47.6&-&-\cr
        AlignGAN\cite{wang2019rgb}&57.9&53.6&56.3&53.4\cr\hline     
        cm-SSFT (Ours)&{\bf 72.3}&{\bf 72.9}&{\bf 71.0}&{\bf 71.7}\cr\hline
    \end{tabular}
    %\vspace{-0.3cm}
\end{table}

\subsection{Ablation study}
In this subsection, we study the effectiveness of each component of the proposed algorithm. 
%Our ablation study includes: (1). Parts ablation: demonstration of the effectiveness of each proposed modules. The experiments are evaluated on RegDB. (2). Feature visualization to represent what the shared and specific features learn, respectively. (3). Different alternative methods for the real application. In this part, we will discuss some unsatisfied cases of our model in the real application and give some effective compromise solutions. Some additional experiments to show the effectiveness of these methods can be seen in the supplementary.  
\begin{table}[tp]
    \centering
    \fontsize{7}{8}\selectfont
    \caption{Ablation study on RegDB.}
    \label{tab:ablation}
    \begin{tabular}{M{0.1cm}|M{0.35cm}M{0.35cm}M{0.35cm}M{0.35cm}M{0.35cm}M{0.35cm}|M{0.35cm}M{0.35cm}|M{0.3cm}M{0.3cm}}
        \hline
        &ShL&SpL&SaS&MoA&PA&RE&ShT&SpT&r1&mAP\\\hline
       1&\checkmark&-&-&-&-&-&-&-&42.4&45.0\\
       2&\checkmark&\checkmark&-&-&-&-&-&-&48.1&49.3\\ 
       3&\checkmark&\checkmark&\checkmark&-&-&-&-&-&52.3&53.1\\\hline
       4&\checkmark&\checkmark&\checkmark&\checkmark&-&-&-&-&56.1&57.2\\
       5&\checkmark&\checkmark&\checkmark&\checkmark&\checkmark&-&-&-&58.7&57.9\\
       6&\checkmark&\checkmark&\checkmark&\checkmark&\checkmark&\checkmark&-&-&60.3&59.4\\\hline
       7&\checkmark&\checkmark&\checkmark&-&-&-&\checkmark&\checkmark&60.8&60.1\\
       8&\checkmark&\checkmark&\checkmark&\checkmark&-&-&\checkmark&\checkmark&67.5&67.6\\
       9&\checkmark&\checkmark&\checkmark&\checkmark&\checkmark&-&\checkmark&\checkmark&71.1&71.2\\\hline
       10&\checkmark&\checkmark&\checkmark&\checkmark&\checkmark&\checkmark&\checkmark&-&65.8&66.1\\
       11&\checkmark&\checkmark&\checkmark&\checkmark&\checkmark&\checkmark&-&\checkmark&64.9&65.3\\
       12&\checkmark&\checkmark&\checkmark&\checkmark&\checkmark&\checkmark&\checkmark&\checkmark&72.3&72.9\\\hline
    \end{tabular}
    %\vspace{-0.3cm}
\end{table}

%\subsubsection{Effectiveness of each modules}
{\bf Effectiveness of structure of feature extractor.} We first evaluate how much improvement can be made by the structure of feature extractor. %In this setting, we do not use feature transfer as well as the complementary learning for fair and clear comparison. 
We ablate the specific feature extraction stream and evaluate the performance of the shared features only to see the influence. The results are shown in the 1$^{\text{st}}$ and 2$^{\text{nd}}$ row of Table~\ref{tab:ablation}, represented as ShL (shared feature learning) and SpL (specific feature learning). The specific streams can bring about 5.7\% increment of Rank-1 accuracy because they can back-propagate modality-specific gradients to the low-level feature maps. We also test the influences casued by separating streams at shallow layers. The result in 3$^{\text{rd}}$ (SaS: Separating at Shallow) shows that it can make bring 4.2\% gains for the more dicriminative features.  

%We aims to analysis the influences of each module in our model. First, for testing the effectiveness of the structure of feature extractor, we ablate the specific stream and evaluate performances of the shared features to see the influences. The results are shown in 1st-2nd lines of the Table \ref{tab:ablation} (represents as ShL (shared feature learning) and ShL+SpL(specific feature learning)). The specific stream can bring about 5.7\% increment because the specific streams in our feature extractor can back-propagate modality-specific gradients to the low-level feature maps, which can make the features more discriminative.

{\bf Influence of complementary learning.} We evaluate the effectiveness of each module in the complementary learning. Since the complementary learning can affect both the features of the feature extractor and SSTN, we design two sets of experiments to observe the impact respectively. The influences on the feature extractor are shown in rows 4$\sim$6 of Table~\ref{tab:ablation}. The results of SSTN are shown in rows 7$\sim$9. We can see that all modules (the modality-adaptation (MoA), the project adversarial (PA) and reconstruction enhancement (RE)) can make both backbone shared features and SSTN features more discriminative. The whole complementary learning scheme can bring about 8\% and 12\% increments for the feature extractor and SSTN, respectively.

{\bf Effectiveness of feature transfer.} We aim to quantify the contribution of the proposed feature transfer strategy. Firstly, we want to know whether the proposed transfer method itself only works on shared features. By comparing row 6 with row 10 (only transfer the shared feature, defined as ShT) in Table~\ref{tab:ablation}, we can see that feature transfer brings in 5.5\% Rank-1 and 6.7\% mAP improvements. 
Secondly, we want to verify whether modality-specific features can positively contribute valuable information to the final representation. According to row 10 and row 12 (transfer both two kinds of features. SpT means transferring specific features.) of Table~\ref{tab:ablation}, we can see that the overall performance gains 6.5\% and 6.8\% increments on Rank-1 and mAP. For further verifying the effectiveness of the specific feature transfer, we also try only transferring the specific features. The results are shown in row 11 and show that only transferring the specific features can also achieve satisfy performances. The feature transfer stage not only contributes an overall 12.0\% Rank-1 and 13.5\% mAP improvements but also verifies that modality-specific features can be well-explored for better re-identification. 

%we want to study whether the performance improvement of SSTN is bring by GCN or bring by the specific information. The results are shown in the line 9 of Table~\ref{tab:ablation}. To analysis this thing, we clear the specific segments in $Z^{(in)}$ and train the model following the unchanged setting. It can be seen that the specific features can gain 6.5\% and 6.8\% on Rank-1 and mAP, which not only shows that our method can utilize specific features effectively but also prove that specific features are useful for cm-ReID. 

\begin{table}[tp]
    \centering
    \fontsize{8}{8}\selectfont
    \caption{Performances without data or network augmentation.}
    \label{tab:tricks}
    \begin{tabular}{M{4cm} M{0.5cm} M{0.5cm} M{0.5cm} M{0.5cm}}
        \hline
        \multirow{2}{*}{Settings}&
        \multicolumn{2}{c}{MM01}&\multicolumn{2}{c}{RegDB}\cr\cline{2-5}&
        r1&mAP&r1&mAP\cr\hline
        SOTA(AlignGAN)'s baseline&29.6&33.0&32.7&34.9\cr
        SOTA(AlignGAN)&42.4&40.7&57.9&53.6\cr\hline
        baseline (wo aug)&25.5&27.2&29.5&30.8\cr
        cm-SSFT (wo aug)&52.4&52.1&62.2&63.0\cr\hline
        baseline (w aug)&38.2&39.8&42.4&45.0\cr
        cm-SSFT (w aug)&61.6&63.2&72.3&72.9\cr\hline
    \end{tabular}
    %\vspace{-0.3cm}
\end{table}

\begin{table}[!t]
    \centering
    \fontsize{8}{8}\selectfont
    \caption{Performances comparison with single query.}
    \label{tab:qsize}
    \begin{tabular}{M{2cm} M{0.3cm} M{0.3cm} M{0.3cm} M{0.3cm} M{0.3cm} M{0.3cm} M{0.3cm} M{0.3cm} M{0.3cm} M{0.3cm} M{0.3cm} M{0.3cm} M{0.3cm} M{0.3cm} M{0.3cm} M{0.3cm} M{0.3cm}}
        \hline
        \multirow{3}{*}{Method}&
        \multicolumn{4}{c}{MM01}&\multicolumn{4}{c}{RegDB}\cr\cline{2-17} &
        \multicolumn{2}{c}{S-shot}&\multicolumn{2}{c}{M-shot}&
        \multicolumn{2}{c}{\emph{V-T}}&\multicolumn{2}{c}{\emph{T-V}}\cr\cline{2-17} &
        r1&mAP&r1&mAP&r1&mAP&r1&mAP\cr\hline
        %For real (no tricks)&40.2&45.6&48.9&49.3\cr
        %SOTA(AlignGAN) &47.7&54.1&57.4&59.1&65.4&65.6&63.8&64.2\cr
        %Baseline&38.2&39.8&44.6&31.7&42.4&45.0&39.8&41.2\cr
        Single query&47.7&54.1&57.4&59.1&65.4&65.6&63.8&64.2\cr
        All queries&61.6&63.2&63.4&62.0&72.3&72.9&71.0&71.7\cr\hline
    \end{tabular}
    %\vspace{-0.3cm}
\end{table}
{\bf Influence of data and network augmentation.} For fair comparison, we also give results without random-erasing in data augmentation. For each feature block, we also use a commonly used fully-connected layer to replace the BNNeck. The results are shown in Table~\ref{tab:tricks}. It can be seen that, without the augmentation, our baseline is weaker than the baseline of the state-of-the-art (AlignGAN~\cite{wang2019rgb}) method (because we don't use dropout). But our model still can suppress  SOTA  by 10.0\% on Rank-1 and 12.1\% on mAP on the SYSU-MM01 dataset. On the RegDB dataset, our method can suppress 4.3\% on Rank-1 and 9.4\% on mAP. The data and network augmentations can bring about 13\% increments on the backbone and 9\% on our method. Without them, our model still achieves the state-of-the-art performances, proving the effectiveness of our method.
\begin{figure}[!t]
    \centering
    \includegraphics[width=0.9\linewidth]{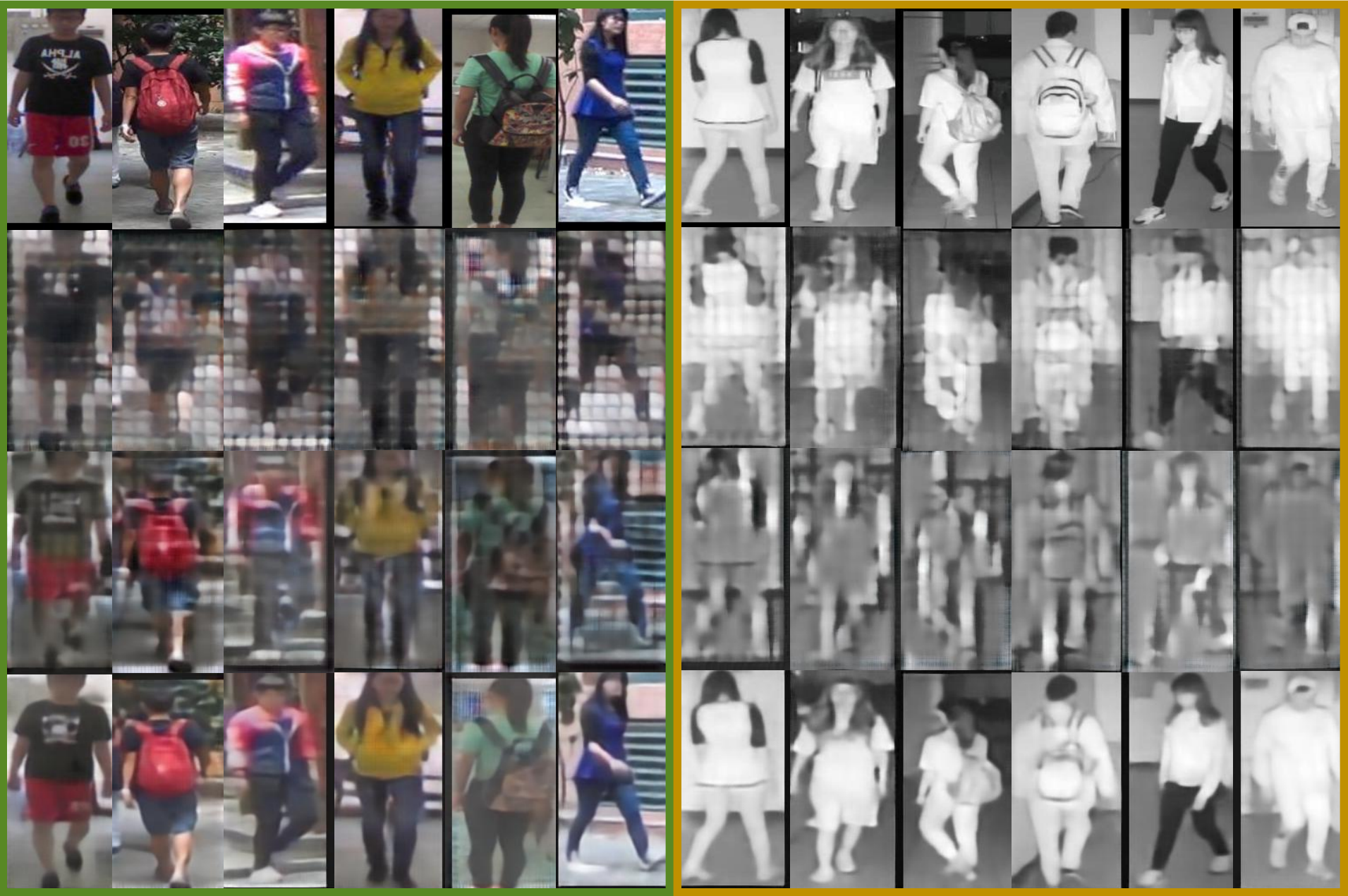}
    \caption{Reconstruction examples. The $1^{st}$ to $4^{th}$ rows correspond to original images, shared feature reconstructions, specific reconstructions and all feature reconstruction results, respectively.} %% label for entire figure
    \label{fig:recon}
    %\vspace{-0.3cm}
\end{figure}

\subsection{Visualization of shared and specific features.}
We take advantage of the reconstruction decoder to visualize the information of the modality shared and specific features. We remove $P^m$ and $H^m$ in Eq.\eqref{eq:decoder} to observe the changes in the reconstructed images, respectively. The outputs are shown in Figure~\ref{fig:recon}. We can see that shared feature reconstruction results are different and visually complementary to the specific features. For RGB images, the shared features contain less color information which is found in the images reconstructed by RGB-specific features. The specific features carried more color information but are less smooth. For infrared images, we can also observe that the specific features are different from the shared features. The combination of two kinds of features produces high-quality images. This proves that the shared and specific features produced by our feature extractor are complementary with each other. %Besides, only using one kind of feature always suffers the blocking artifact due to the vast amount of information loss.

\subsection{Application in real scenarios}
The SSTN in our cm-SSFT passes information between different modality samples. Every sample fuses the information from its inter-modality and intra-modality $k$ near neighbors. Such setting hypothesizes that other query samples are treated as the auxiliary set. However, in some real application scenarios, there may be no or only a few auxiliary dates. In order to prove that our method is not limited in the experimental environments with some strong hypothesis, we show how to apply cm-SSFT to such single query scenarios, which also achieves state-of-the-art performances. We train the cm-SSFT algorithm exactly the same as illustrated in this paper. While in the testing stage, the SSTN only propagates information between only one query image with the gallery images. We slightly stabilize the affinity model $A$ as follows: 
\begin{small}\begin{equation}
%\label{normalize}
%\begin{aligned}
  Z=\begin{pmat}[{}]
            z^q \cr\-
            Z^G \cr
    \end{pmat},
      A=\begin{pmat}[{|}]
            k\cdot A^{q,q} & \mathcal{T}(A^{q,G},k) \cr\-
            k\cdot A^{G,q} & \mathcal{T}(A^{G,G},k) \cr
    \end{pmat}.
%\end{aligned}
 \label{eq:affinity}
\end{equation}\end{small}
%\begin{small}\begin{equation}
%    S=\begin{pmat}[{|}]
%            k\cdot s^{q,q} & \mathcal{T}(S^{q,G},k) \cr\-
%            k\cdot S^{G,q} & \mathcal{T}(S^{G,G},k) \cr
%    \end{pmat}.
%    \label{eq:affinity}
%\end{equation}\end{small}

It can be seen that we amplify $k$ times the left column blocks of the affinity matrix, which is to balance the information of the two modalities. The experiments are shown in Table~\ref{tab:qsize}. The performance has dropped compared with all queries due to inadequate intra-modality specific information compensation. But our method still achieves better performances than state-of-the-arts and our baseline.
% Moreover, the other reason is the gap between the training and the testing. The training schedule guide the model to utilize all two modality-specific information. But in testing, there is only a little information of the query modality. Motivated by this observation, we propose the Dual Relation Learning (DRL, detatils can be seen in the supplementary) to alleviate the stage gap between the training and the testing. The results show that, with the DRL, the model can get better performance, which is not far away from the standard one. 
\begin{figure}[!t]
    \centering
    \includegraphics[width=0.9\linewidth]{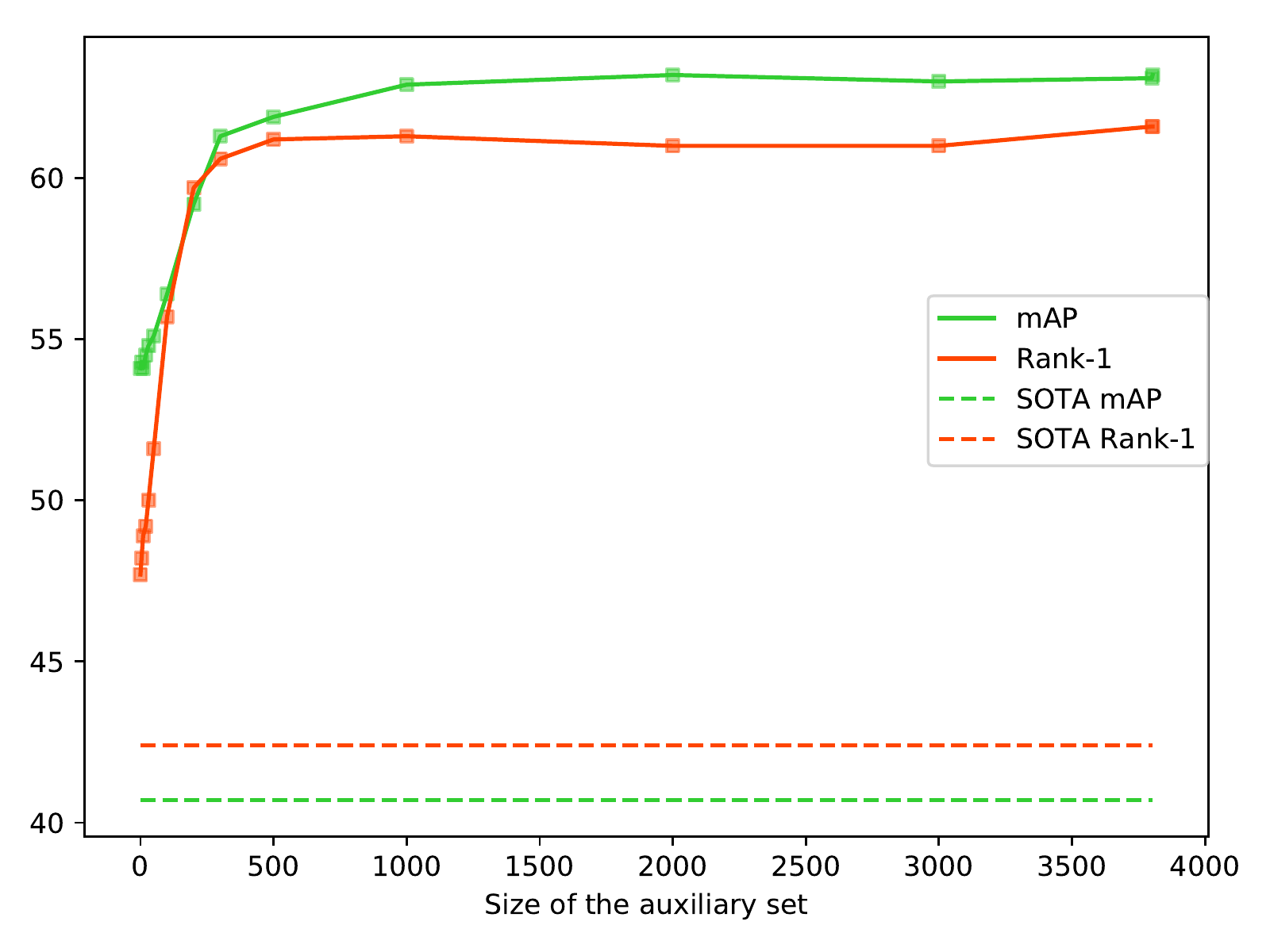}
    \caption{Influence of number of queries. Dashed lines correspond to the SOTA method. Solid lines correspond to ours. } %% label for entire figure
    \label{fig:q_size}
    %\vspace{-0.3cm}
\end{figure}

Besides, we also test the influence of the auxiliary set. The experiments are run on MM01 dataset for its large query set. We randomly sample $n$ images from the query sets and watch the performance changing. For a specific $n$, we run 10 times to get the average performance. $n$ is ranging from 1 (single query) to all query size. The results are shown in Figure~\ref{fig:q_size}. We can see that with the size of the auxiliary set growing, the performance saturates quickly. 
%Even on the single query case, our method still outperforms SOTA by 13.4\%, 12\% mAP and 5.3\%, 7.5\% rank-1 accuracy on SYSU-MM01 and RegDB. 

\section{Conclusion}
In this paper, we proposed a cross-modality shared-specific feature transfer algorithm for cross-modality person ReID, which can utilize the specific features ignored by conventionally shared feature learning. It propagates information among and across modalities, which not only compensates for the lacking specific information but also enhances the overall discriminative. We also proposed a complementary learning strategy to learn self-discriminate and complementary feature. Extensive experiments validate the superior performance of the proposed algorithm, as well as the effectiveness of each component of the algorithm. 

\section{Acknowledgement}
This work was supported by the Fundamental Research Funds for the Central Universities (WK2100330002, WK3480000005) and the Major Scientific Research Project of Zhejiang Lab (No.2019DB0ZX01). 
%-------------------------------------------------------------------------

%-------------------------------------------------------------------------

%-------------------------------------------------------------------------

%------------------------------------------------------------------------

\end{document}